\documentclass{article}

\usepackage{amsmath,amsfonts,bm}

\def\eqref#1{equation~\ref{#1}}

\def\1{\bm{1}}

\newcommand{\test}{\mathcal{D_{\mathrm{test}}}}

\DeclareMathAlphabet{\mathsfit}{\encodingdefault}{\sfdefault}{m}{sl}
\SetMathAlphabet{\mathsfit}{bold}{\encodingdefault}{\sfdefault}{bx}{n}

\DeclareMathOperator*{\argmin}{arg\,min}

\usepackage[utf8]{inputenc} 
\usepackage[T1]{fontenc}    
\usepackage{hyperref}       
\usepackage{url}            
\usepackage{booktabs}       
\usepackage{amsfonts}       
\usepackage{nicefrac}       
\usepackage{amsmath,amsthm}
\newtheorem{theorem}{Theorem}
\newtheorem{corollary}{Corollary}
\usepackage{graphicx} 
\usepackage{wrapfig}
\usepackage{enumitem}

\usepackage[accepted]{icml2024}
\usepackage[capitalize,noabbrev]{cleveref}

\newcommand{\fsig}{\ensuremath{f_{\text{sig}}}}
\newcommand{\fsq}{\ensuremath{f_{\text{sq}}}}
\newcommand{\fsqrt}{\ensuremath{f_{\text{sqrt}}}}

\newcommand{\dof}{dof}
\newcommand{\dofs}{\dof}

\newcommand{\x}{\boldsymbol{x}}
\newcommand{\w}{\boldsymbol{w}}
\newcommand{\fnet}{f_{\text{\normalfont net}}}
\newcommand{\fsmooth}{f_{\text{\normalfont smooth}}}
\newcommand{\f}{f}
\newcommand{\X}{\mathcal{X}}
\newcommand{\Etr}{E_{\text{train}}}

\newcommand{\smooth}{smooth-mini-max}

\newcommand{\MM}{MM}
\newcommand{\SMM}{SMM}
\newcommand{\HLL}{HLL}
\newcommand{\HLLs}{HLL$^{\text{s}}$}
\newcommand{\HLLl}{HLL$^{\text{l}}$}
\newcommand{\LMN}{LMN}
\newcommand{\LMNs}{LMN$^{\text{s}}$}
\newcommand{\LMNl}{LMN$^{\text{l}}$}

\newcommand{\XGs}{XG$^{\text{s}}$}
\newcommand{\XGl}{XG$^{\text{l}}$}
\newcommand{\XGsval}{XG$^{\text{s}}_{\text{val}}$}
\newcommand{\XGlval}{XG$^{\text{l}}_{\text{val}}$}
\newcommand{\Iso}{Iso}
\newcommand{\XG}{XG}
\newcommand{\XGval}{XG$_{\text{val}}$}

\renewcommand{\test}{\ensuremath{\mathcal{D}_{\text{test}}}}
\newcommand{\sigdif}{$^*$}
\newcommand{\low}[1]{\textbf{#1}}

\newcommand{\ysmm}{y_{\text{SMM}}}
\newcommand{\gsmm}{g_{\text{SMM}}}
\newcommand{\ntrain}{N_{\text{train}}}
\newcommand{\ntest}{N_{\text{test}}}
\newcommand{\dtrain}{\ensuremath{\mathcal{D}_{\text{train}}}}
\newcommand{\dtest}{\ensuremath{\mathcal{D}_{\text{test}}}}
\newcommand{\lse}{\ensuremath{\operatorname{LSE}}}

\icmltitlerunning{Smooth Min-Max Monotonic Networks}
\begin{document}
\twocolumn[
\icmltitle{Smooth Min-Max Monotonic Networks}

\begin{icmlauthorlist}
\icmlauthor{Christian Igel}{yyy}
\end{icmlauthorlist}
\icmlaffiliation{yyy}{Department of Computer Science, University of Copenhagen, Copenhagen, Denmark}
\icmlcorrespondingauthor{Christan Igel}{igel@diku.dk}
\icmlkeywords{monotonic neural networks, deep learning, fairness}

\vskip 0.3in
]

%
\printAffiliationsAndNotice{}

\begin{abstract}\noindent
Monotonicity constraints are powerful regularizers in statistical modelling. They can support fairness in computer-aided decision making and increase plausibility in data-driven scientific models. The seminal min-max (MM) neural network architecture ensures monotonicity, but often gets stuck in undesired local optima during training because of partial derivatives of the MM nonlinearities being zero. We propose a simple modification of the MM network using strictly-increasing smooth minimum and maximum functions that alleviates this problem. The resulting smooth min-max (SMM) network module inherits the asymptotic approximation properties from the MM architecture. It can be used within larger deep learning systems trained end-to-end. The SMM module is conceptually simple and computationally less demanding than state-of-the-art neural networks for monotonic modelling. Our experiments show that this does not come with a loss in generalization performance compared to alternative neural and non-neural approaches.
\end{abstract}

\section{Introduction}
\setlength\intextsep{0ex}
\begin{figure}[hbt!]
\includegraphics[width=\columnwidth, trim={2em .4em 4em 4em},clip]{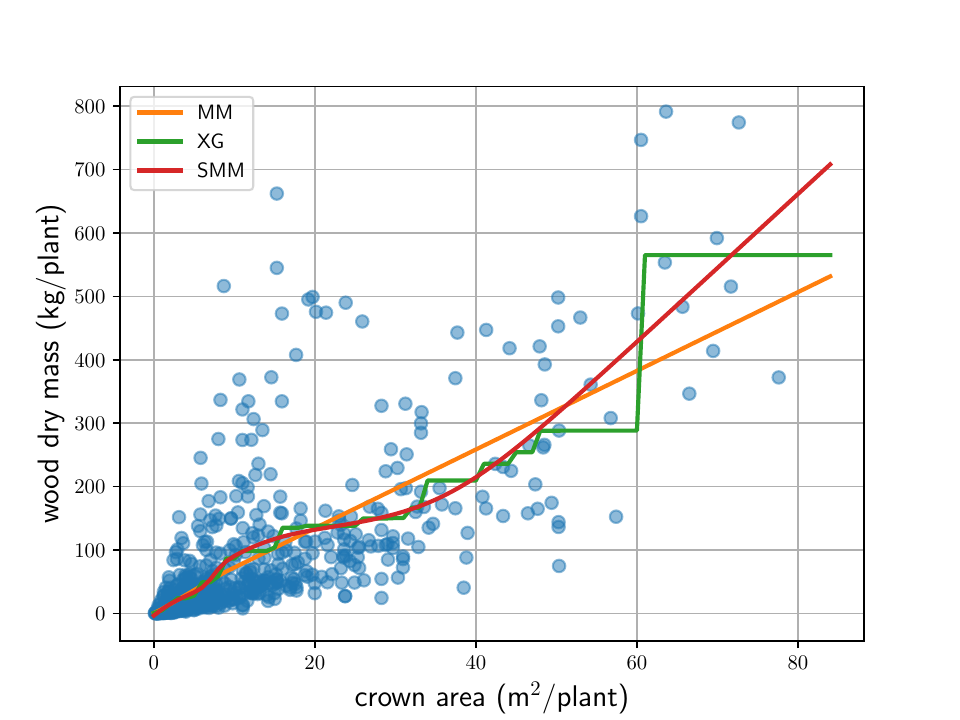} 
\caption{Learning an allometric equation from data with an original min-max network (MM), XGBoost (XG) and a smooth min-max network (SMM), here estimating wood dry mass (and thereby stored carbon) from tree crown area \citep{hiernaux:23,tucker:23}.\label{fig:allo}}
\end{figure}
In many data-driven modelling tasks we have a priori knowledge that the output is monotonic, that is, non-increasing or non-decreasing, in some of the input variables.
This knowledge can act as a regularizer, and often monotonicity is a strict constraint for ensuring the plausibility and therefore acceptance of the resulting model. We are particularly interested in  monotonicity constraints when learning bio- and geophysical models from noisy observations, see  Figure~\ref{fig:allo}. Examples from finance, medicine and engineering are given, for instance, by \citet{daniels:2010}, see also the review by \citet{cano:19}. 
Monotonicity constraints can incorporate 
ethical principles into data-driven models and improve their fairness \citep[e.g., see][]{cole2019avoiding,wang2020deontological}.

\setlength\intextsep{12.0pt plus 2.0pt minus 2.0pt}

Work on monotonic neural networks was pioneered by the min-max (MM) architecture proposed by \citet{sill:97}, which is simple, elegant, and able to asymptotically approximate any monotone target function by a piecewise linear neural network model.
However, learning an MM network, which can be done by unconstrained gradient-based optimization, often does not lead to satisfactory results. Thus,  a variety of alternative approaches were proposed,  which are much more complex than an MM network module (for recent examples see \citealp{fard:16,you:17,gupta2019incorporate,yanagisawa2022hierarchical,sivaraman2020counterexample,liu2020certified}, and \citealp{nolte2022expressive}). We argue that the main problem when training an MM network are partial derivatives being zero because of the maximum and minimum computations. This leads to large parts of the MM network being silent, that is, most parameters of the network do not contribute to computing the model output at all, and therefore the MM network underfits the training data with a very coarse piecewise linear approximation.  We alleviate this issue by  replacing the maximum and minimum by smooth \emph{and monotone} counterparts. The resulting neural network module is referred to as \emph{smooth min-max} (SMM) and exhibits the following properties:
\begin{itemize}[leftmargin=*]
\item The SMM network inherits the asymptotic approximation properties of the min-max architecture, but does not suffer from large parts of the network not being used after training. 
\item The SMM module can be used within a larger deep learning system and be trained 
end-to-end using unconstrained gradient-based optimization in contrast to standard isotonic regression and (boosted) decision trees.
\item The SMM module is simple and does not suffer from the curse of dimensionality  when the number of constrained inputs increases, in contrast to lattice based approaches. 
\item The function learned by SMM networks is smooth in contrast to isotonic regression, linearly interpolating lattices, and boosted decision trees.
\item Our experiments show that the advantages of SMM do not come with a loss in performance. In  experiments on elementary target functions, SMM compared favorably with min-max networks, isotonic regression, XGBoost, expressive Lipschitz monotonic
networks, and hierarchical lattice layers; and \SMM{} also worked well on partial monotone real-world benchmark problems.
\end{itemize}
We would like to stress that the smoothness property is not just a technical detail. It influences how training data are inter- and extrapolated, and smoothness can be important for scientific plausibility.
Figure~\ref{fig:allo} shows an example where an allometric equation is learned from noisy observations using the powerful XGBoost \citep{Chen:2016} as well as  simple MM and \SMM{} layers. In this example, the output (wood dry mass) should be continuously increasing with the input (tree crown area). The MM layer collapses to a linaet function.
Both XGBoost and the \SMM{} layer  give good fits in terms of mean squared error, neither the  staircase shape nor the constant extrapolation of the tree-based model are scientifically plausible.

The next section will present  basic theoretical results on neural networks with positive weights and the MM architecture as well as a brief overview of interesting alternative neural and non-neural approaches to monotonic modelling. After that, Section~\ref{sec:smm} will introduce the SMM module and show that it inherits the  asymptotic approximation properties from MM networks. Section \ref{sec:experiments} will present an empirical evaluation of the SMM module with a clear focus on the monotonic modelling capabilities in comparison to alternative neural and non-neural approaches before we conclude in Section~\ref{sec:conclusion}.

\section{Background}

A function $f(\x)$ depending on $\x=(x_1,\dots,x_d)^{\text{T}}\in\mathbb R^d$
is non-decreasing in variable $x_i$ if $x_i'\ge x_i$ implies $f(x_1,\dots,x_{i-1},x'_i,x_{i+1}, \dots,x_d) \ge f(x_1,\dots,x_{i-1},x_i,x_{i+1}, \dots,x_d)$; being non-increasing is defined accordingly. A function is called monotonic if it is non-increasing or non-decreasing in all $d$ variables. Without loss of generality, we assume that monotonic functions are non-decreasing in all $d$ variables (if the function is supposed to be non-increasing in a variable $x_i$ we simply negate the variable and consider $-x_i$). We address the task of inferring a monotonous model from noisy measurements. For regression we are given samples $\mathcal{D}_{\text{train}}=\{(\x_1,y_1),\dots,(\x_n, y_n)\}$ where $y_i = f(\x_i) + \varepsilon_i$ with $f$ being monotonic and $\varepsilon_i$ being a realization of a random variable with zero mean. Because of the random noise,  $\mathcal{D}_{\text{train}}$ is not necessarily a monotonic data set, which implies that interpolation does in general not solve the task.

\subsection{Neural Networks with Positive Weights}
\paragraph{Basic theoretical results.} 
A common way to enforce monotonicity of canonical neural networks is to restrict the weights to be non-negative.
If the activation functions are monotonic, then a network with non-negative weights is also monotonic \citep{archer93,sill:97,daniels:2010}.
However, 
this does not ensure that the
resulting network class can approximate any monotonous function arbitrarily well. 
If the activation functions of the hidden neurons are standard sigmoids (logistic/Fermi functions) and the output neuron is linear (e.g., the activation function is the identity), then a neural network with positive weights and at most $d$ layers can approximate 
any continuous function mapping from a compact subset of $\mathbb R^d$ to $\mathbb R$ arbitrarily well \citep[][Theorem 3.1]{daniels:2010}. Interesting recent theoretical work by \citet{mikulincer:22} shows that with Heaviside step activation functions  the above result can be achieved with four layers for non-negative inputs (their interpolation results assume monotone data and are therefore not applicable to the general case of noisy data).
However, if the activation functions in the hidden layers are convex, such as the popular (leaky) ReLU and ELU \citep{nair2010rectified,maas2013rectifier,clevert2015fast} activation functions, then a canonical neural network with positive weights is a combination of convex functions and as such convex, and accordingly one can find a non-convex monotonic function that cannot be approximated within an a priori fixed additive error \citep[][Lemma 1]{mikulincer:22}. 

\paragraph{Min-max networks.}
\begin{figure}[hbt]
    \centering
    \includegraphics[width=0.4\textwidth]{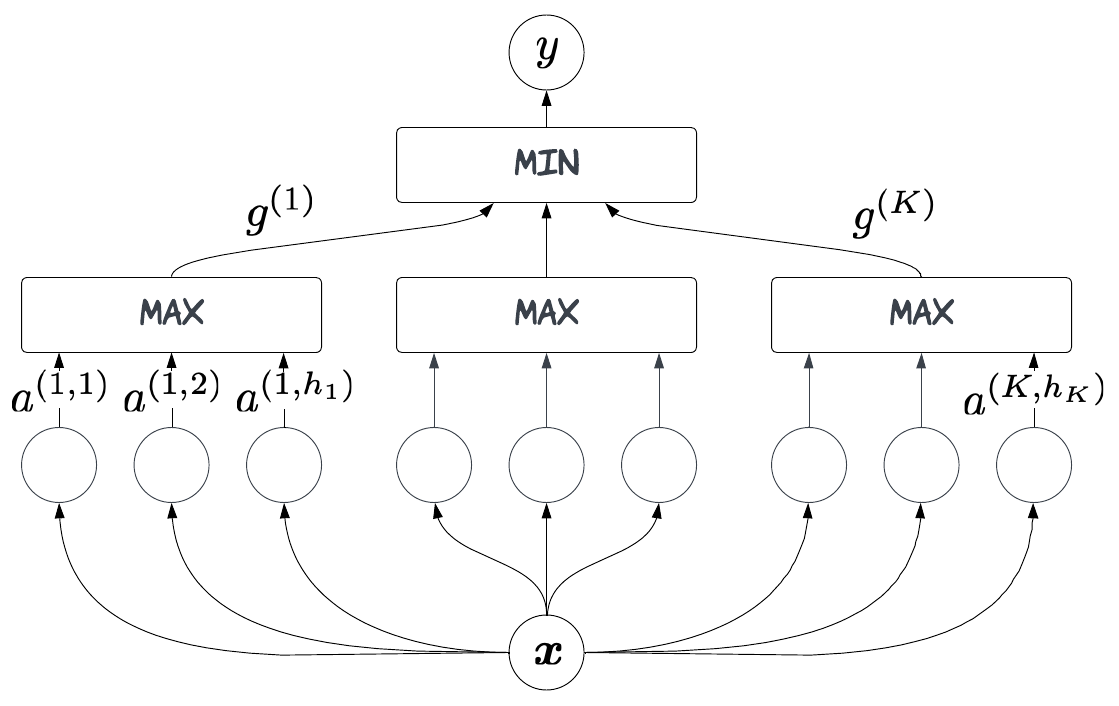}
    \caption{Schema of a min-max module.}
    \label{fig:minmax}
\end{figure}
Min-max (MM) networks as proposed by \citet{sill:97} are a concave combination -- taking the minimum -- of convex combinations --  taking the maximum -- of monotone linear functions, where the monotonicity is ensured by positive weights, see Figure~\ref{fig:minmax}.
The architecture comprises $K$ groups of linear neurons, where, following the original notation, the number of neurons in group $k$ is denoted by $h_k$.  
Given an input $\x\in\mathbb R^d$, neuron $j$ in group $k$ computes
\begin{equation}\label{eq:activation}
    a^{(k,j)}(\x) =  \w^{(k,j)}\cdot \x - b^{(k,j)}
\end{equation}
with weights $\w^{(k,j)}\in (\mathbb{R}^+_0)^d$ and bias $b^{(k,j)}\in\mathbb R$.
Then all $h_k$ outputs within a group $k$ are combined via 
\begin{equation}\label{eq:g}
    g^{(k)}(\x) = \max_{1\le j\le h_k} a^{(k,j)}(\x)
\end{equation}
and the output of the network is given by
\begin{equation}\label{eq:y}
    y(\x)=\min_{1\le k\le K} g^{(k)}(\x)\enspace.
\end{equation}
For classification tasks, $y$ can be interpreted as the logit.
To ensure positivity of weights during unconstrained optimization, we encode each weight $w^{(k,j)}_i$ by 
an unconstrained parameter $z^{(k,j)}_i$, where $w^{(k,j)}_i=\exp\big(z^{(k,j)}_i\big)$ \citep{sill:97} or
$w^{(k,j)}_i$
results from squaring \citep{daniels:2010} or applying the 
exponential linear function \citep{cole2019avoiding} to $z^{(k,j)}_i$.
The order of the minimum and maximum computations can be reversed 
\citep{daniels:2010}.
The convex combination of concave functions gives the following asymptotic approximation capability:
\begin{theorem}[\citealp{sill:97,daniels:2010}]
Let $\f(\x)$ be any continuous, bounded monotonic function with bounded partial derivatives, mapping $[0,1]^d$ to $\mathbb R$. Then there exists a function $\fnet(\x)$ which can be implemented by a monotonic network such that $|\f(\x)-\fnet(\x)|<\epsilon$ for any $\epsilon>0$ and any $\x\in[0,1]^d$.\label{thm:sill}
\end{theorem}

\subsection{Related Work}

\paragraph{Lattice layers.}
Neural networks with lattice layers constitute a state-of-the-art approach for incorporating monotonicity constraints \citep{fard:16,you:17,gupta2019incorporate,yanagisawa2022hierarchical}.
A lattice layer defines a hypercube with $L^d$ vertices. The integer hyperparameter $L>1$ defines the granularity of the hypercube and $d$ is the input dimensionality, which is replaced by the number of input features with monotonicity constraints  in \emph{hierarchical lattice layers} (HLLs, \citealp{yanagisawa2022hierarchical}).
In contast to the original lattice approaches, a HLL can be trained by  unconstrained gradient-based-optimization.
The $L^d$ scaling of the number of parameters is a limiting factor. For larger $d$, the task has to be broken down using an ensemble of several lattice layers, each handling fewer constraints \citep{fard:16}.


\paragraph{Certified monotonic neural networks.}
A computationally very expensive approach to monotonic modelling is to train standard piece-wise linear (ReLU) networks and to ensure monotonicity afterwards.
\citet{liu2020certified} propose to train with heuristic regularization that favours monotonicity. After training, it is checked by solving a MILP (mixed integer linear program) if the network fulfills all constraints. If not, the training is repeated with stronger regularization. 
\citet{sivaraman2020counterexample} suggest to adjust the output of the trained network to ensure monotonicity. This requires solving an SMT (satisfiability modulo theories, a generalization of SAT) problem for each prediction.

\paragraph{Lipschitz monotonic networks.}
\citet{nolte2022expressive} have proposed  Lipschitz monotonic networks (LMNs). The idea of LMNs is to ensure that 
a base model is  $\lambda$-Lipschitz   with respect to the
$L^1$-norm and then to add $\lambda x_i$ to the model for each  constrained input $i$.
LMNs are smooth and can be trained end-to-end.
The LMN approach requires choosing the Lipschitz constant $\lambda$.
To enforce the Lipschitz property of neural models, normalization of the weight matrices is added.
However, to ensure that the networks can approximate any monotonic Lipschitz bounded function,
one has to additionally use special activation functions to prevent ``gradient attenuation'' (in the experiments by \citeauthor{nolte2022expressive} the GroupSort activation function was used), see also \citet{pmlr-v97-anil19a}. 
This approximation result is slightly weaker than Theorem~\ref{thm:sill} in the sense that the choice of $\lambda$ constrains the class of functions the LMN can approximate.

\paragraph{Constrained monotonic neural networks.} \citet{runje23}  have been recently proposed constrained monotonic neural networks (CMNNs). 
To ensure monotonicity, these networks constrain the weights to be positive. In order to be able to approximate any monotonic function, the neurons within a CMNN layer use three different activation functions. 
Given some zero-centered, non-decreasing, convex, lower-bounded activation function (e.g., the ReLU), two additional activation functions are constructed: the corresponding concave function resulting from reflecting the graph both vertically and horizontally 
(similar to the work by \citealp{eides:18}) and a bounded function constructed from the other two other functions. The CMNN layers can be trained using unconstrained gradient-based optimization.
The approach is simple and elegant and enjoys asymptotic approximation properties similar to Theorem~\ref{thm:sill}.

\paragraph{Non-neural approaches.}
There are many  approaches to monotonic prediction not based on neural networks, we refer to \citet{cano:19} for a survey.
We would like to highlight isotonic regression (\Iso), which is often used for classifier calibration \citep[e.g., see][]{niculescu2005predicting}. In its canonical form  (e.g., see \citealp{best1990active} and \citealp{de2009isotone}),  \Iso{} fits a piece-wise constant function to the data and is restricted to univariate problems.
The popular XGBoost gradient boosting library \citep{Chen:2016} also supports monotonicity constraints.
XGBoost incrementally learns an ensemble of decision trees; accordingly, the resulting regression function is piece-wise constant.

\section{Smooth Monotonic Networks}\label{sec:smm}
We now introduce the smooth min-max (SMM) network module, which addresses problems of the original MM architecture. 
The latter often performs worse than alternative approaches both in terms of training and test error, and  the outcome of the training process strongly depends on the initialization. 
%
Even if an MM architecture has enough neurons to be able to 
approximate the underlying target functions well (see Theorem \ref{thm:sill}), the neural network parameters realizing this approximation may not be found by the (gradient-based) learning process.
When using MM modules in practice, they often underfit the training data and seem to approximate the data using a piecewise linear model with very few pieces –– much less than the number of neurons.
This observation is  empirically studied in Section~\ref{sec:active}. 
We say that neuron  $j^*$ in group $k^*$ in an MM unit is \emph{active} for an input $\x$, if 
$k^*= \argmin_{1\le k\le K} g^{(k)}(\x)$ and $j^*= \max_{1\le j\le h_k} a^{(k,j)}(\x)$.
A neuron is \emph{silent} over a set of inputs $\mathcal X\subset \mathbb R^d$ if it is not active for any $\x\in\mathcal X$. If neuron  $j$ in group $k$ is silent over all inputs from some training set $\dtrain$, we have 
$\nicefrac{\partial y}{\partial a^{(k,j)}}(\x) = 0$ for all $\x\in\mathcal X$. Once a neuron is silent over the training data, which can easily be the case directly after initialization or happen during training, there is a high chance that gradient-based training will not lead to the neuron becoming active. 
Indeed, our experiments in Section~\ref{sec:active} show that only a small fraction of the 
neurons in an MM module are active when the trained model is evaluated on test data.

The  problem of silent neurons and the lack of smoothness can be addressed by replacing 
the minimum and maximum operation in the MM architecture by smooth counterparts. Not every  approximation to the maximum/minimum function is suitable, it has to preserve monotonicity,
need to work for positive and negative arguments,
should have a bounded approximation error which can be controlled (see Corollary~\ref{cor}), 
should be smooth, and need to be computable efficiently without numerical problems.
The LogSumExp function has all these properties.
Let  $x_1,\dots,x_n \in \mathbb R$.
We define the scaled LogSumExp function with scaling parameter $\beta>0$
as
\begin{multline}\label{eq:lse}
\lse_\beta(x_1,\dots,x_n) = \frac{1}{\beta} \log\sum_{i=1}^n \exp(\beta x_i) \\= \frac{1}{\beta} \left( c + \log\sum_{i=1}^n \exp(\beta x_i - c) \right) \enspace,
\end{multline}
where the constant $c$ can be freely chosen to increase numerical stability, in particular as $c=\max_{1\le i\le n} x_i$.
The functions $\lse_\beta(\X)$ and $\lse_{-\beta}(\X)$ are smooth and monotone increasing in $x_1,\dots,x_n$.
It holds: 
\begin{equation}
    \max_{1\le i\le n} x_i < \lse_\beta(x_1,\dots,  x_n)
    \le \max_{1\le i\le n} x_i + \frac{1}{\beta} \ln(n)
    \end{equation}
\begin{equation}
    \min_{1\le i\le n} x_i - \frac{1}{\beta} \ln(n) \le  \lse_{-\beta} (x_1,\dots, x_n)
    < \min_{1\le i\le n} x_i
\end{equation}
The proposed SMM module is identical to an MM module, except that \cref{eq:g}
and \cref{eq:y} are replaced by 
\begin{align}\label{eq:gbeta}
    \gsmm^{(k)}(\x) &= \lse_\beta\big(a^{(k,1)}(\x),\dots, a^{(k,h_k)}(\x)\big)
\text{ and }\\\label{eq:ybeta}
    \ysmm(\x) &=\lse_{-\beta}\big(\gsmm^{(1)}(\x), \dots, \gsmm^{(K)}(\x)\big) \enspace.
\end{align}
We treat $\beta$, properly encoded to ensure positvity, as an additional learnable parameter. 
Thus, the number of 
 parameters of an SMM module is $1+ (d+1)\sum_{k=1}^K h_k$.
If the target function is known to be (strictly) concave, we can set $K=1$ and $h_1>1$; if it is known to be convex, we set $K>1$ and can set $h_k=1$ for all $k$. The default choice is $K=h_1=h_2=\dots= h_K$.

We can rewrite the above definition such that the $\beta$ parameter appears only once, rescaling the final output.
The  $\beta$ factors acting on the $a^{(k,j)}(\x)$ can be absorbed by the parameters $\w^{(k,j)}$ and $b^{(k,j)}$ 
(which could be considered by the initializing of these parameters). The outer $\beta$ factors in \cref{eq:gbeta} and the inner $\beta$ factors in \cref{eq:ybeta} cancel. Thus we get the equivalent simpler definition
\begin{align}\label{eq:gsimple}
    \gsmm^{(k)}(\x) &= \lse_1\big(a^{(k,1)}(\x),\dots, a^{(k,h_k)}(\x)\big)
\text{ and }\\\label{eq:ysimple}
    \ysmm(\x) &=\beta \lse_{-1}\big(\gsmm^{(1)}(\x), \dots, \gsmm^{(K)}(\x)\big) \enspace,
\end{align}
in which the role of $\beta$ is just a final linear rescaling. 

\subsection{Approximation Properties}
The SMM inherits the approximation properties from the MM, e.g.:
\begin{corollary}\label{cor}
Let $\f(\x)$ be any continuous, bounded monotonic function with bounded partial derivatives, mapping $[0,1]^d$ to $\mathbb R$. Then there exists a function $\fsmooth(\x)$ which can be implemented by a smooth monotonic network such that $|\f(\x)-\fsmooth(\x)|<\epsilon$ for any $\epsilon>0$ and any $\x\in[0,1]^d$.
\end{corollary}%
\begin{proof}
Let $\epsilon=\gamma+\delta$ with $\gamma>0$ and $\delta>0$.
From Theorem~\ref{thm:sill} we know that there exists an MM network $\fnet$ with  $|\f(\x)-\fnet(\x)|<\gamma$. Let $\fsmooth$ 
be the smooth monotonic neural network as defined by 
\cref{eq:gbeta} and \cref{eq:ybeta}  with the same weights and bias parameters as $\fnet$.
Let $H=\max_{h=1}^K h_k$.  For all $\x$ and groups $k$ we have 
\begin{align}
    \gsmm^{(k)}(\x) &=  \lse_\beta\big(a^{(k,1)}(\x),\dots, a^{(k,h_k)}(\x)\big)\notag\\
&\le \max_{1\le j\le h_k} a^{(k,j)}(\x) + \frac{1}{\beta} \ln(h_k) \notag\\ 
&\le g^{(k)}(\x) + \frac{1}{\beta} \ln(H)\enspace. 
\end{align}
Thus, also $\ysmm(\x)  \le y(\x) + \frac{1}{\beta} \ln(H)$. 
Similarly, we have 
\begin{align}
\ysmm(\x) &=\lse_{-\beta}\big(\gsmm^{(1)}(\x), \dots, \gsmm^{(K)}(\x)\big) \notag\\
&\ge \lse_{-\beta}\big(g^{(1)}(\x), \dots, g^{(K)}(\x)\big) \notag\\
&\ge  \min_{1\le k\le K} g^{(k)}(\x) - \frac{1}{\beta} \ln(K) \notag\\ 
&= y(\x) -\frac{1}{\beta}\ln(K)\enspace.
\end{align}
Thus, setting $\beta=\delta^{-1}\ln\max(K,H)$ ensures for all $\x$ that $|\fnet(\x)-\fsmooth(\x)|\le\delta$ and therefore $|\f(\x)-\fsmooth(\x)|< \gamma+\delta = \epsilon$.
\end{proof}

\subsection{Partial Monotonic \SMM}
Let $\X$ be a subset of variables from $\{x_1,\dots,x_d\}$.
Then a function is partial monotonic in $\X$ 
if it is monotonic in all $x_i\in\X$.
The min-max and \smooth{} modules are partial monotonic in $\X$ 
if the positivity constraint is imposed for weights connecting to $x_i\in\X$ \citep{daniels:2010}; the other weights can vary freely.
However, more general module architectures are possible. Let us split the input vector into $(\x^{\text{c}}, \x^{\text{u}})$, where $\x_{\text{c}}$ comprises all $\X$ and $\x_{\text{u}}$ the remaining $x_i\not\in\X$. Let $\Psi^{(k,j)}:\mathbb R^{d-|\X|}\to  (\mathbb{R}^+_0)^{|\X|}$ and $\Phi^{(k,j)}:\mathbb R^{d-|\X|}\to \mathbb R^{l^{(k,j)}}$ for some integer $l^{(k,j)}$ denote neural subnetworks for each neuron $j=1,\dots,h_k$ in each group $k=1,\dots,K$ (which may share weights). 
Then replacing \cref{eq:activation} by $a^{(k,j)}(\x) =   \w^{(k,j)}\cdot \x + \Psi^{(k,j)}(\x_{\text{u}}) \cdot \x_{\text{c}} + \w^{(k,j)}_{\text{u}}\cdot \Phi(\x_{\text{u}}) - b^{(k,j)}$ with $\w^{(k,j)}_{\text{u}}\in\mathbb{R}^{l^{(k,j)}}$ and $\forall m\in\mathcal{X}:w^{(k,j)}_m\ge 0$ preserves the constraints.

\section{Experiments}\label{sec:experiments}
We empirically compared different monotonic modelling approaches on well-understood benchmark functions. We also present results for various partial monotonic real-world data sets.\footnote{All experiments, plots, tables, and statistics can be reproduced using the  source code available from \url{https://github.com/christian-igel/SMM}.}
As in related studies, the results on the partial monotonic real-world data reflect the general inductive bias of the overall system architecture, not only the performance of the network modules handling monotonicity constraints; this bears the risk that the processing of the unconstrained features occludes the monotonic modelling performance. 

In our experiments, we assumed that we do not have any prior knowledge about the shape of the target function and set $K=h_1=h_2,=\dots=h_K=6$. 
%
To avoid hyperparameter overfitting, we used the these hyperparameters for the SMM modules in \emph{all} experiments.
 We use the exponential encoding to ensure positive weights. The weight parameters 
${z}_i^{(k,j)}$ and the bias parameters were randomly initialized by samples from a 
 Gaussian distribution with zero mean and unit variance truncated to $[-2,2]$. We also used exponential encoding of $\beta$ and initialize $\ln \beta$ with $-1$.

We compared against isotonic regression (\Iso) as implemented in the Scikit-learn library \citep{scikit-learn} and XGBoost (\XG, \citealp{Chen:2016}).
As initial experiments showed a tendency of \XG\ to overfit, we evaluated \XG\ with and without early-stopping.
We considered hierarchical lattice layers (HLL) as a state-of-the-art representative 
of lattice-based approaches using the well-documented 
implementation made available by the authors\footnote{\url{https://ibm.github.io/pmlayer}}. For a comparison of HLL with other lattice models we refer to \citet{yanagisawa2022hierarchical}.
Furthermore, we applied LMNs using the implementation by \citeauthor{nolte2022expressive}.\footnote{\url{https://github.com/niklasnolte/MonotonicNetworks}}
For our new experiments, we adopted the basic architecture used in the \emph{ChestXRay} experiments 
by
\citet{nolte2022expressive} with two hidden layers and Lipschitz parameter one. The number of neurons in the hidden layers is determined by a width parameter. In each experiment, we considered two model sizes. The width parameter should be even, and we picked the width such that the model size (in degrees of freedom) of the small \LMNs\ is smaller or equal to the size of the corresponding SMM. The larger \LMNl\ used a width parameter increased by two compared to \LMNs. The resulting model sizes  embrace the corresponding SMM model size, see next section and Table~\ref{tab:ucidata} and  Table~\ref{tab:multitraintime} in the appendix.
In our experiments, the neural network models SMM, MM,  HLL, and LMN were trained by the same unconstrained iterative gradient-based optimization procedure.

\subsection{Univariate Modelling}
We considered  three simple basic univariate functions on $[0,1]$, the convex $\fsq(x)=x^2$ ,  the concave $\fsqrt(x)=\sqrt{x}$, and the scaled and shifted logistic function $\fsig=(1 + \exp(-10(x-\nicefrac{1}{2}))^{-1}$; see also the work by \citet{yanagisawa2022hierarchical} for experiments on $\fsq$ and $\fsqrt$.
For each experimental setting, $T=21$ independent trials were conducted.
For each trial, the $\ntrain{}=100$ training data points \dtrain{} were generated by randomly sampling inputs from the domain. Mean-free Gaussian noise with standard deviation $\sigma=0.01$
was added to target outputs (i.e., the training data  were typically not monotone, in contrast to, e.g., the setting considered by \citealp{mikulincer:22}). 
The test data \dtest{} were  noise-free evaluations of $\ntest=1000$  evenly spaced inputs covering the input domain.

We compared SMM, MM,  HLL, LMN as well as isotonic regression (Iso) and XGBoost (XG) with and without early-stopping.
For $K=6$, the MM and SMM modules have 72 and 73 trainable parameters, respectively.
We matched the degrees of freedom and set the number of vertices in the HLL to 73; \LMNs\ and \LMNl\ had
width parameters 6 and 8 resulting in 61 and 97 trainable parameters, respectively.
We  set the number of estimators in XGBoost to $n_{\text{trees}}=73$ and $n_{\text{trees}}=35$ (as the behavior was similar, we report only the results for $n_{\text{trees}}=73$ in the following); for all other hyperparameters the default values were used. When using XGBoost with early-stopping, referred to as \XGval, we used 25\,\% of the training data for validation and set the number of early-stopping rounds to $\lfloor\nicefrac{n_{\text{trees}}}{10}\rfloor$. 
The isotonic regression baseline requires specifying the range  of the target functions, and also HLL presumes a codomain  of $[0,1]$. This is useful prior information not available to the other methods, in particular as some of the training labels may lie outside this range because of the added noise. 
We evaluated the methods by their mean-squared error (MSE). Details of the gradient-based optimization are given in Appendix~\ref{app:opt}.


The test and training results of the experiments on the univariate functions are summarized in Table~\ref{tab:uniresults} and Table~\ref{tab:uniresultstrain}, respectively. The distribution of the results is visualized in Figure~\ref{fig:bar1D0.01}.
In all experiments SMM gave the smallest median test error, and all  differences between SMM and the other methods were statistically significant (paired two-sided Wilcoxon test, $p<0.001$).
The lower training errors of \XG\ and \Iso\ indicate overfitting.
However, in our experimental setup, early-stopping in \XGval{} did not improve the overall performance.
The lattice layer performed better than XGBoost.
\SMM{} was statistically significantly better than \HLL\ and both \LMN\ variants;` the latter did not perform well in this experimental setup.
Figure~\ref{fig:singletrial} depicts the results of a random trial, showing  the different ways the  models extra- and interpolate.
 
\paragraph{Silent neurons.}
Overall, SMM clearly outperformed MM. The variance of the MM learning processes was significantly higher, see Figure~\ref{fig:bar1D0.01}.  This can be attributed to the problem of silent neurons; the MM training got stuck in undesired local minima.
When looking at  the $3\cdot 21=63$ trials on the univariate test functions after training, the maximum number of  \MM\ neurons at least once active over the test data set was as low as 5 out of $36$; the mean number of active neurons was 2.8.
On average 3.7 neurons in a network were active directly after initialization, that is, the training typically decreased the number of active neurons.\footnote{Before developing the SMM, we tried to solve the problem of silent neurons by improving the initialization, however, without success.}
For SMM, we inspected the sum of the test predictions $\sum_{(x,y)\in\dtest} \ysmm(x)$ after training.
We counted for how many neurons  both partial derivatives of this sum
  w.r.t.{} the neuron's parameters were zero, which could happen for numerical reasons.
This was rarely the case. On average more than 31 neurons were active after training using this notion of activity and never less than 14. Detailed results for MM and SMM are given in Table~\ref{tab:uniactive} in the appendix.
\label{sec:active}

\paragraph{Robustness.}
\emph{After these experiments,} we evaluated the robustness of the \SMM{} results 
for different choices of initial $\ln\beta\in\{-3, -2, -1, 0, 1\}$ and $K=h_k\in\{2, 4, 6, 8\}$. The results are shown in Table~\ref{tab:unihyper} in the appendix. Our default choice of $\beta=-1$ with $K=6$ was suboptimal in all cases.
We used \cref{eq:gsimple} and \cref{eq:ysimple} without changing the initialization of the weights and biases, and as expected the choice of the initial $\beta$ had little influence on the performance.
These results show the robustness of the \SMM{} approach and  that $\beta$ does not introduce a sensitive hyperparameter but  just one additional model weight.


\begin{table*}[ht]
\caption{Median test errors on univariate (top) and mutivariate (bottom) tasks based on 21 trials per experimental setting. A star indicates that the difference on the test data in comparison to \SMM{} is statistically significant (paired two-sided Wilcoxon test, $p<0.001$). The mean-squared error (MSE) values are multiplied by $10^3$.\label{tab:uniresults}\label{tab:multi}}
\begin{center}
\begin{tabular}{@{}lllllllll@{}}
\toprule
& {\MM} & \SMM & {\XG} & \XGval & {\Iso} & {\HLL} & \LMNs & \LMNl\\
 \midrule
\fsq & 0.10\sigdif & \low{0.01} & 0.14\sigdif & 0.18\sigdif & 0.04\sigdif & 0.04\sigdif & 0.37\sigdif & 0.09\sigdif \\
\fsqrt & 0.32\sigdif & \low{0.02} & 0.14\sigdif & 0.20\sigdif & 0.06\sigdif & 0.06\sigdif & 0.28\sigdif & 0.27\sigdif \\
\fsig & 0.22\sigdif & \low{0.01} & 0.13\sigdif & 0.17\sigdif & 0.04\sigdif & 0.04\sigdif & 0.25\sigdif & 0.26\sigdif \\
\bottomrule
\end{tabular}

\medskip
\begin{tabular}{@{}lrrrrrrrrr@{}}
\toprule
 & \SMM & \XGs & \XGsval & \XGl & \XGlval& \HLLs & \HLLl & \LMNs & \LMNl\\
 \midrule
$d=2$ & \low{0.00} & 0.23\sigdif & 0.26\sigdif & 0.23\sigdif & 0.26\sigdif & 0.03\sigdif & 0.03\sigdif & 0.07\sigdif & 0.03\sigdif \\
$d=4$ & \low{0.01} & 0.66\sigdif & 0.76\sigdif & 0.66\sigdif & 0.76\sigdif & 0.03\sigdif & 0.08\sigdif & 0.29\sigdif & 0.06\sigdif \\
$d=6$ & \low{0.02} & 0.74\sigdif & 0.82\sigdif & 0.74\sigdif & 0.82\sigdif & 0.10\phantom{\sigdif} & 0.13\sigdif & 0.07\phantom{\sigdif} & 0.07\sigdif \\
 \bottomrule
\end{tabular}
\end{center}
\end{table*}

\subsection{Multivariate Functions}
We evaluated \SMM, \XG,  \HLL\, and \LMN\ on multivariate monotone target functions.
The original \MM\ was dropped because of the previous results, \Iso\ because the considered algorithm does not extend to multiple dimensions in a canonical way (the Scikit-learn implementation  only supports univariate tasks).
We considered three input dimensionalities $d\in\{2,4,6\}$.
In each trial, we randomly constructed a function.
Each function mapped a $[0,1]^d$ input to its polynomial features up to degree 2 and computed a weighted sum of these features, where for each function the weights were drawn independently uniformly from $[0,1]$ and then normalized by the sum of the weights. 
For example, for $d=2$ we had $(x_1, x_2)^{\text{T}}\mapsto (w_1 + w_2x_1+ w_3x_2+ w_4x_1^2+ w_5x_2^2+ w_6x_1x_2)\cdot \left(\sum_{i=1}^6w_i\right)^{-1}$ with $w_1,\dots,w_6\sim U(0,1)$. We uniformly sampled $\ntrain=500$ and $\ntest=1000$ training and test inputs from $[0,1]^d$, and noise was added as above.

For $K=6$, the dimensionalities result in 109, 181, and 253 learnable parameters for the SMM.
The number of learnable parameters for HLL is given by the $L^d$ vertices in the lattice. In each trial, we considered 
two lattice sizes.
For \HLLs{}, we set  $L$ to 10, 3, and 2 for $d$ equal to 2, 4, and 6, respectively; for \HLLl{} we increased $L$ to 11, 4, and 3, respectively. We also considered two \LMN{} architectures. For both \LMN{} and \HLL{} 
the  smaller network had fewer and the larger had more degrees of freedom than the corresponding SMM,
see Table~\ref{tab:multitraintime} in the appendix.
We ran XGBoost with $n_{\text{trees}}=100$ (\XGs) and $n_{\text{trees}}=200$ (\XGl), with and without early-stopping.

The test error results of $T=21$ trials are summarized in Table~\ref{tab:multi}. The corresponding training errors are shown in Table~\ref{tab:multitrain} in the appendix. The boxplot Figure~\ref{fig:barmulti0.01}  in the appendix visualizes the results.
The newly proposed \SMM\ statistically significantly outperformed all other algorithms in all settings, except \HLLs\ and \LMNs\ for $d=6$ where the lower errors reached by \SMM\ are not significant.
Using early stopping did not improve the XGBoost results in our setting, and doubling the number of trees did not have a considerable effect on training and test errors.
We also measured the neural network training times for 1000 iterations, see Table~\ref{tab:multitraintime} in the Appendix~\ref{app:add}. \HLLs{} 
was more than an order of magnitude slower  than \LMNs\ and the fastest method \SMM. 

\subsection{UCI Partial Monotone Functions}\label{sec:uci}
\newcommand{\energyOne}{Energy Efficiency $Y_1$}
\newcommand{\energyTwo}{Energy Efficiency $Y_2$}
\newcommand{\qsar}{QSAR Aquatic Toxicity}
\newcommand{\concrete}{Concrete Compressive Strength}

\renewcommand{\energyOne}{Energy  $Y_1$}
\renewcommand{\energyTwo}{Energy  $Y_2$}
\renewcommand{\qsar}{QSAR}
\renewcommand{\concrete}{Concrete}
\newcommand{\energy}{Energy} 

\newcommand{\MLPSMM}{SMM$_{64}$}
As a proof of concept, we considered modelling partial monotone functions on real-world data sets from the UCI benchmark repository \citep{Dua:2019}.
Details about the experiments are provided in Appendix~\ref{app:uci}.
We took all regression tasks and constraints from the first group of benchmark functions considered by \citet{yanagisawa2022hierarchical}. The input dimensionality $d$ and number of constraints $|\mathcal{X}|$ were $d=8$ and $|\mathcal{X}|=3$  for
the \emph{Energy Efficiency} data  \citep{tsanas2012accurate} (with two regression targets $Y_1$ and $Y_2$), $d=6$ and $|\mathcal{X}|=2$ for the 
\emph{\qsar} data \citep{cassotti2015similarity}, and 
$d=8$ and $|\mathcal{X}|=1$ for \emph{\concrete} \citep{yeh1998modeling}.
We performed 5-fold cross-validation. From each  fold available for training, 25\,\%  were used as a validation data set for early-stopping and final model selection, giving a 60:20:20 split in accordance with \citet{yanagisawa2022hierarchical}.  
In the partial monotone setting, HLL internally uses an auxiliary neural network. 
We used a network with a single hidden layer with 64 neurons, which gave better results than the larger default network.
We considered SMM  with unrestricted weights for the unconstrained inputs. 
We also added an auxiliary network. The 
\MLPSMM{} model computes  $a^{(k,j)}(\x) = \w^{(k,j)} \cdot \x + 
 \Phi(\x_{\text{u}})
- b^{(k,j)}$, where $\Phi:\mathbb{R}^{d-|\mathcal{X}|}\to \mathbb{R} $ is a neural network with 64 hidden units processing the unconstrained  inputs, see Appendix~\ref{app:uci} for details. Similar to \HLL, we incorporate the knowledge about the targets being in $[0,1]$ by applying a standard sigmoid to the activation of the output neuron. 
We present \XG\ results for $n_{\text{trees}}=100$, increasing the number of trees to $n_{\text{trees}}=500$ did not yield superior results.

The mean cross-validation test error is shown in Table~\ref{tab:uci}.
\MLPSMM{} performed best for one task, \XG\ in the others.
\MLPSMM{} had the lowest CV test error of the neural network approaches on the two \energy\ tasks, and the larger \LMN{} on \qsar\ and on \concrete.


\newcommand{\ntrees}{\ensuremath{n_{\text{trees}}}}
\begin{table*}[ht!]
\caption{Results on partial monotone UCI tasks, cross-validation error averaged over the MSE of 5 folds. The MSE is multiplied by 100. The dof columns give the numbers of trainable parameters, $\ntrees$ the maximum number of estimators in XGBoost.\label{tab:uci}}
\begin{center}
\begin{tabular}{@{}lrrrrrrrrrrrr@{}}
\toprule
& \multicolumn{2}{c}{ \MLPSMM } & \multicolumn{2}{c}{ \SMM } & \multicolumn{2}{c}{ \XG } & \multicolumn{2}{c}{ \HLL } & \multicolumn{2}{c}{ \LMNs } & \multicolumn{2}{c}{ \LMNl } \\
  & \test & \dof & \test & \dof &  \test &\ntrees &  \test & \dof &  \test & \dof  &  \test & \dof \\
 \midrule
\energyOne &  \low{0.14} &  774 &  0.25 &  325 &  0.22 &  100 &  0.45 &  2139 &  0.27 &  727 &  0.22 &  841 \\
\energyTwo &  0.24 &  774 &  0.61 &  325 &  \low{0.11} &  100 &  0.29 &  2139 &  0.44 &  727 &  0.34 &  841 \\
\qsar &  1.03 &  638 &  1.02 &  253 &  \low{0.98} &  100 &  0.99 &  905 &  1.01 &  581 &  0.99 &  683 \\
\concrete &  1.78 &  902 &  1.79 &  325 &  \low{1.71} &  100 &  4.59 &  707 &  2.20 &  841 &  1.71 &  963 \\
\bottomrule
\end{tabular}
\end{center}
\end{table*}

\newcommand{\lmnds}[1]{#1}
\begin{table*}[ht!]
\caption{Comparison on common benchmark functions.
The results for counterexample-guided learning of monotonic neural networks (COMET),  Lipschitz monotonic networks (LMNs) and certified monotonic neural networks (Certified) are taken from \citet{nolte2022expressive}, the results 
for XGBoost (XG), constrained monotonic neural networks (CMNN), and lattice ensembles (Crystals, \citealp{fard:16}) from \citet{runje23}. The SMM experiments used the code from \citet{nolte2022expressive} and exactly their experimental setup (three trials, etc.), see caption of their Table~1.
Accuracies and corresponding standard deviations are given in percent. \MLPSMM\ sig.~refers to the architecture with sigmoidal output activation.\label{tab:expressive}}
\begin{center}
\begin{tabular}{@{}l@{}ccccccc@{}}
\toprule
\multicolumn{1}{@{}c}{} & \lmnds{COMPAS} & \lmnds{BlogFeedback} & \lmnds{LoanDefaulter} & \multicolumn{2}{@{}c}{\lmnds{ChestXRay}} & \lmnds{Heart Disease} & \lmnds{Auto MPG} \\
Method & $\uparrow\uparrow$ Test Acc & $\downarrow\downarrow$ RMSE  & $\uparrow\uparrow$ Test Acc & $\uparrow\uparrow$ Test Acc & $\uparrow\uparrow$ Test Acc & $\uparrow\uparrow$ Test Acc & $\downarrow\downarrow$ MSE  \\
&&&& pretrained & end-to-end\\
\midrule
Certified &  68.8$\,\pm\,$0.2   & 0.158$\,\pm\,$0.001 &  65.2$\,\pm\,$0.1  & 62.3$\,\pm\,$0.2 & 66.3$\,\pm\,$1.0 \\
LMN &  69.3$\,\pm\,$0.1  & 0.160$\,\pm\,$0.001 &  65.44$\,\pm\,$0.03 &  67.6$\,\pm\,$0.6 &  {70.0$\,\pm\,$1.4}   &   89.6$\,\pm\,$1.9\phantom{0}  & 7.58$\,\pm\,$1.2\phantom{0} \\
LMN mini & & 0.155$\,\pm\,$0.001 &  65.28$\,\pm\,$0.01 \\
COMET &&&&&& \phantom{.0}86$\,\pm\,$3\phantom{.00}  & 8.81$\,\pm\,$1.81  \\
\hline
Crystal	&66.3$\,\pm\,$0.1& 0.164$\,\pm\,$0.002 &	\phantom{0}65.0$\,\pm\,$0.1\phantom{0}	\\
CMNN	&69.2$\,\pm\,$0.2 &	{0.156$\,\pm\,$0.001}	& \phantom{0}65.3$\,\pm\,$0.01	& & & \phantom{.0}89$\,\pm\,$0\phantom{.00}	& 8.37$\,\pm\,$0.08\\
XG & 68.5$\,\pm\,$0.1& 0.176$\,\pm\,$0.005 &	\phantom{0}63.7$\,\pm\,$0.1\phantom{0}	\\
\hline
\MLPSMM & \textbf{69.5$\,\pm\,$0.1} & 0.192$\,\pm\,$0.002 &  65.41$\,\pm\,0.03$ & \textbf{67.9$\,\pm\,$0.4} &  \textbf{70.1$\,\pm\,$1.2} & 88.5$\,\pm\,$1.0\phantom{0}   & \textbf{7.51$\,\pm\,$1.6\phantom{0}}  \\
\MLPSMM\ mini & & \textbf{0.154$\,\pm\,$0.000\makebox[0pt][l]{4}} &  \textbf{65.47$\,\pm\,$0.00\makebox[0pt][l]{3}} \\ 
\MLPSMM\ sig. &&&&&& \textbf{91.3$\,\pm\,$1.89} \\
\bottomrule
\end{tabular}
\end{center}
\end{table*}
\subsection{Comparison with Recently Published Results}
The question arises how our approach compares to the results on larger real-world data sets presented by \citet{nolte2022expressive}.
Thanks to \citeauthor{nolte2022expressive} who make their code for their experiments available,\footnote{\url{https://github.com/niklasnolte/MonotonicNetworks}} we could evaluate 
\SMM{} \emph{exactly} as in their work.
Additionally, we compared to the corresponding results reported by \citet{runje23}.
We employed the  \MLPSMM{} model already used in Section~\ref{sec:uci}.
As done by \citet{nolte2022expressive},
we conducted only three trials, not enough to establish that the observed differences are  statistically significant.
Note that the evaluation procedure implemented by \citet{nolte2022expressive} assumes an oracle identifying the network with the lowest test error during training (i.e., the results in Table~\ref{tab:expressive} are not unbiased estimates of generalization performance).
It has to be stressed that the \LMN{} results presented 
by \citet{nolte2022expressive} 
were produced using different network architectures and different hyperparameters of the learning algorithm for the different tasks. In contrast, we achieved our results using a single  architecture which was not tuned for the tasks. We also used exactly the same number of training steps, we only adjusted the learning rates.
For the \emph{Heart Disease} task, we also provide the results when adding an additional sigmoid to the output and a slightly longer training time.

We added our experimental results to the values for LMNs, certified monotonic neural networks \citep{liu2020certified}  and counterexample-guided learning of monotonic neural networks \citep[COMET,][]{sivaraman2020counterexample}
as given by \citet{nolte2022expressive} and to the results for XGBoost (XG), constrained monotonic neural networks (CMNN),  and lattice ensembles (Crystals, \citealp{fard:16}) from \citet{runje23}.
SMM models gave better results in all of the benchmarks. For \emph{BlogFeedback} we profited from the feature selection used by \citet{nolte2022expressive}. For Heart Disease, the architecture with the additional output sigmoid gave the best results  (if we use the same  number of training iterations the average result equals the 89.6 reported for \LMN).

\section{Conclusions}\label{sec:conclusion}
The smooth min-max (\SMM) module is a simple, efficient, theoretically sound,  and -- as we  would argue -- very elegant way to ensure monotonicity. 
The experiments confirmed our hypothesis that the pioneering min-max (\MM) architecture suffers from 
silent neurons. This issue is addressed by the \SMM{}, which 
is the main reason why the proposed approach achieves state-of-the-art performance.
In  light of our  results, many neural network approaches for modelling monotonic functions  appear overly complex, both in terms of algorithmic description length and especially computational complexity.
For example, lattice-based approaches suffer from the exponential increase in the number of trainable parameters with increasing dimensionality, and other approaches rely on solving SMT and MILP problems, which are  typically NP-hard.
The SMM is designed to be a module usable in a larger learning system that is trained end-to-end.
From the methods considered in this study,  MM, \HLL, CMNN, and \LMN{} have this property, and we regard SMM as a drop-in replacement for those. 

Which of the monotonic regression methods considered in this study results in a better generalization performance is of course task dependent. The different models have different inductive biases.
All artificial benchmark  functions considered in our experiments were smooth, matching the -- rather general and highly relevant -- application domain the SMM module was developed for. 
The monotonicity constraints of SMM act as a strong regularizer, and overfitting was no problem in our experiments. The \SMM{} approach does not add hyperparameters to \MM{}. All SMM experiments were performed with a single hyperparameter setting for the architecture. This shows the robustness of the method.
We regard the way SMM networks inter- and extrapolate (see Figure~\ref{fig:allo} and Figure~\ref{fig:singletrial}) as a big advantage over \XG, \HLL, and \Iso{} for the type of scientific modelling tasks that motivated our work.
\LMN{}s and CMNNs   share many of the desirable properties of \SMM{}s. 
\LMN{}s require imposing an upper bound on the Lipschitz constant of the network.
Such a bound can act as a regularizer  and supports theoretical analysis of the neural network.
Thus, if such a bound is  desired anyway, the \LMN{} approach is a convenient way to additionally ensure monotonicity.
However, a wrongly chosen bound can limit the approximation capabilities. 
The current asymptotic approximation results are less general for \LMN{}s compared to CMNNs and \SMM{}s.
CMNNs appear to perform very similar to \SMM{} and seem to be a comparable alternative.
However, our experiments show that there are no reasons to prefer \LMN{}s or CMNNs over \SMM{}s because of generalization performance and efficiency.

In summary, \SMM{} modules provide an efficient way to ensure monotonicity. They inherit  the simplicity and the asymptotic approximation guarantees from of the original min-max approach and performed well in our experimental evaluation 
without architecture and hyperparameter tuning.

\section*{Acknowledgements}
I thank the Villum Foundation for their support through the project \emph{Deep Learning and Remote Sensing for Unlocking Global Ecosystem Resource Dynamics (DeReEco)} and the Pioneer Centre for AI, DNRF grant number P1.

\section*{Impact Statement}
This paper presents work whose goal is to advance the field of Machine Learning (ML). There are many potential societal consequences of our work.
We would like to highlight 
the importance of  monotonicity constraints for the fairness of ML systems.
As for example \citet{wang2020deontological}
 point out,  
 monotonicity can be used to  implement 
 ethical principles and social norms such as ``favor the less fortunate'' and ``do not penalize good attributes.''
\bibliography{smooth}
\bibliographystyle{icml2024}

\newpage
\appendix
\onecolumn


\renewcommand\thefigure{\thesection.\arabic{figure}} 
\renewcommand\thetable{\thesection.\arabic{table}} 
\renewcommand\theequation{\thesection.\arabic{equation}}

\section{Gradient-based Optimization}\label{app:opt}
The neural network models SMM, MM, and HLL were fitted by unconstrained iterative gradient-based optimization of 
the mean-squared error (MSE) on the training data.
We used the Rprop optimization algorithm \citep{riedmiller1993direct,igel:01e}.
On the fully monotone benchmark functions, we did not have a validation data set for stopping the training. 
Instead, we monitored the \emph{training progress over a training strip of length $k$} defined by 
\citet{Prechelt2012} as 
$P_k(t)=10^3\cdot\left(\frac{\sum_{t'=t-k+1}^t \Etr(t')}{k\cdot\min^t_{t'=t-k+1}\Etr(t')}\right)$ for $t\ge k$. Here $t$ denotes the current iteration (epoch) and  $\Etr(t')$ the MSE on training data at iteration $t'$. Training is stopped as soon the progress falls below a certain threshold $\tau$. We used $k=5$ and $\tau=10^{-3}$. This is a very conservative setting which worked well for HLL and was then adopted for all algorithms.

\section{Details on UCI Experiments}\label{app:uci}
The experiments on  partial monotone functions were inspired by \citet{yanagisawa2022hierarchical}.
As briefly discussed in Section~\ref{sec:experiments}, a fair comparison  on complex   partial monotone real-world tasks
is challenging. There is the risk that the performance on the unconstrained features overshadows the processing of the constraint features. Therefore, we did not consider the second group of UCI tasks from the study by \citeauthor{yanagisawa2022hierarchical}, because the fraction of constrained features in these problems is too low -- and we would argue that the low number of constrained features already is an issue for the problems in the first group when evaluating monotone modelling. We selected all regression tasks from the first group, see the overview in Table~\ref{tab:uci}. We used the same  constraints, see Table~\ref{tab:uci}, and normalization to $[0,1]$ of inputs and targets as 
\citet{yanagisawa2022hierarchical}. 

Furthermore, architecture and hyperparameter choices become more important in the UCI experiments compared to the experiments on the comparatively simple benchmark functions.
For partial monotone tasks, the HLL requires an auxiliary neural network. The default network did not give good results in initial experiments, so we replaced it by a network with a single hidden layer with 64 neurons, which performed considerably better.
The lattice sizes of the constrained input features were set to $k=3$.

For a fair comparison, we also added an auxiliary network with 64 neurons to the SMM module. For complex real-world tasks, an isolated SMM module with a single layer  of adaptive weights -- despite the asymptotic approximation results -- is not likely to be the right architecture. 
Thus, we considered SMM modules with a single  neural network  $\Phi:\mathbb R^{d-|\X|}\to \mathbb R^{d}$ with one hidden layer and  compute $a^{(k,j)}(\x) =   \w^{(k,j)}\cdot \x + \Phi(\x_{\text{u}}) - b^{(k,j)}$, where $d$ is the input dimensionality, $\x_{\text{u}}$ are the unconstrained inputs,  $|\mathcal X|$ is  the number of constrained variables, and  $\forall m\in\mathcal{X}:w^{(k,j)}_m\ge 0$, see  end of Section~\ref{sec:smm}. We set the number of hidden neurons of $\Phi$ to 64, so that degrees of freedom are similar to the HLL employed in our experiments.
Also similar to \HLL, we incorporate the knowledge about the targets being in $[0,1]$ by applying a standard sigmoid $\sigma$ to the activation of the output neuron. 
The resulting architecture, which we refer to as \MLPSMM{}, can alternatively be written as as a residual block computing
$\sigma(y(\x) + \Phi(\x_{\text{u}}))$, where $y(\x)$ is the standard \SMM. This may be the simplest way to augment the \SMM{}. 
%

\begin{table*}[ht!]
\caption{UCI regression data sets and constraints as considered by \citet{yanagisawa2022hierarchical}. The input dimensionality is denoted by $d$, the number of data points by $n$. The last five columns give the number of trainable parameters of the models used in the experiments; SMM and \MLPSMM{} denote the smooth min-max network without and with  auxiliary neural network $\Phi$. \label{tab:ucidata}}
\begin{center}
\begin{tabular}{@{}lrrrrrrrr@{}}
\toprule
& $d$ & $n$ & monotone features &\multicolumn{5}{c}{no.~parameters}\\
& & &  & SMM & \MLPSMM{} & \HLL & \LMNs & \LMNl \\
 \midrule
Energy  &  8 & 768  & X3, X5, X7 & 325 & 744 & 2139\ & 727 & 841\\
\qsar &  6 & 908 & MLOGP, SM1\_Dz(Z) &  253 & 638 & 905 & 581 & 683 \\
\concrete & 8 &  1030  & Water & 325 & 902 &  707    &  841 & 963\\
\bottomrule
\end{tabular}
\end{center}
\end{table*}

 We performed 5-fold cross-validation to evaluate the methods.
Each data fold available for training was again split to get a validation data set, giving a 60:20:20 spit into training, validation, and test data as considered by \citet{yanagisawa2022hierarchical}.
We monitored the  MSE on the validation data during training and stored the model with the smallest validation loss.
If the validation error did not decrease for 100 epochs, the training was stopped.


\section{Additional Results}\label{app:add}

\begin{table*}[ht]
\caption{Training errors on univariate tasks.  The mean-squared error (MSE) values are multiplied by $10^3$.\label{tab:uniresultstrain}}
\begin{center}
\begin{tabular}{@{}lllllllll@{}}
\toprule
& {\MM} & \SMM & {\XG} & \XGval & {\Iso} & {\HLL} & \LMNs & \LMNl\\
 \midrule
\fsq & 0.17 & 0.10 & 0.05 & 0.11 & 0.03 & 0.03 & 0.42 & 0.14 \\
\fsqrt & 0.35 & 0.09 & 0.04 & 0.10 & 0.03 & 0.03 & 0.31 & 0.25 \\
\fsig & 0.27 & 0.10 & 0.05 & 0.11 & 0.04 & 0.04 & 0.36 & 0.43 \\
\bottomrule
\end{tabular}
\end{center}
\end{table*}

\begin{figure*}[ht!]
    \centering
    \includegraphics[width=\textwidth]{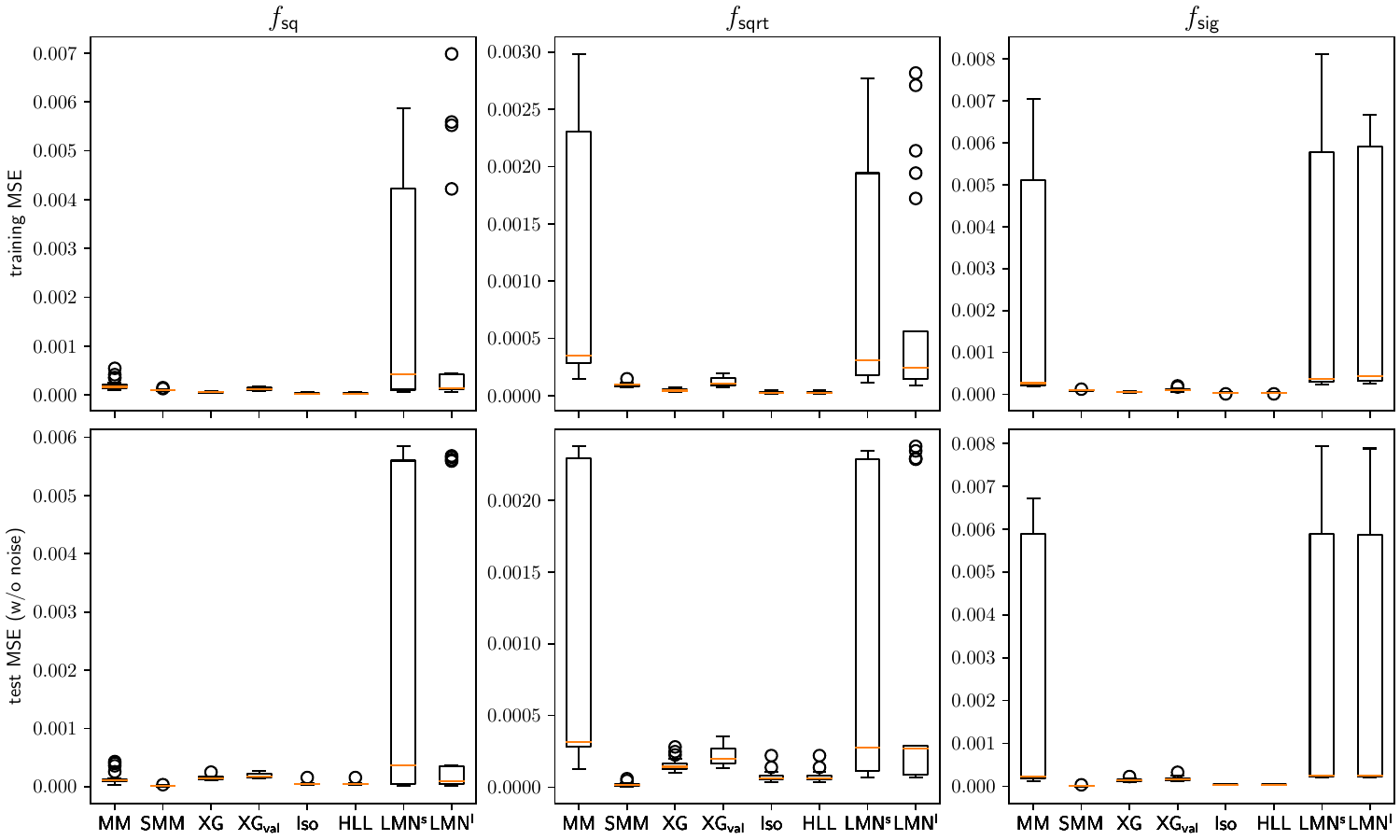}
    \caption{Results on univariate functions based on $T=21$ trials. Depicted are the median, first and third quartile of the MSE (without clipping the outputs to the target function codomain); the  whiskers extend the box by $1\nicefrac{1}{2}$ the inter-quartile range, dots are outliers. Training errors are shown in the top, test errors in the bottom row.}
    \label{fig:bar1D0.01}
\end{figure*}

\begin{figure*}[ht]
    \centering
    \includegraphics[width=\textwidth]{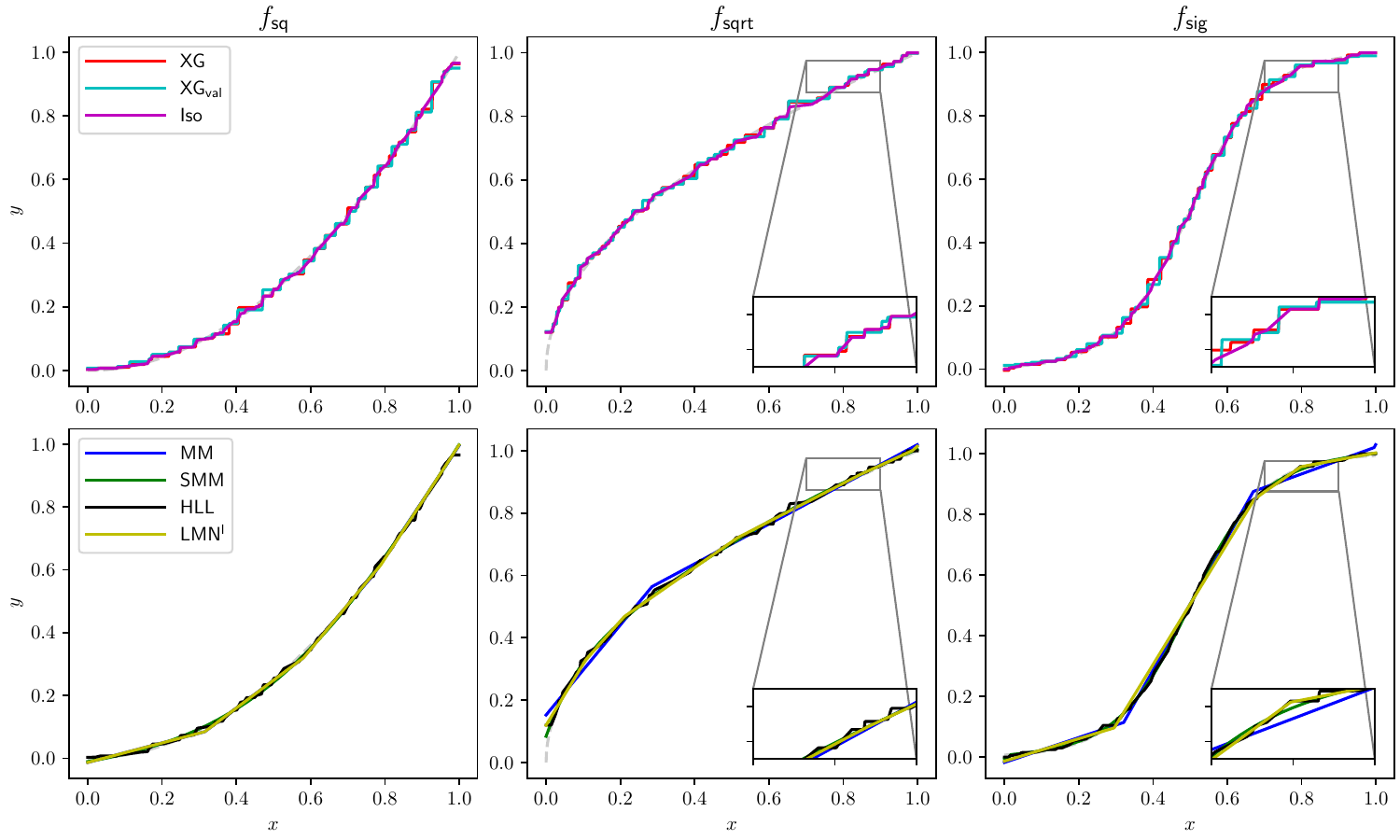}
    \caption{Function approximation results of a single trial (outputs not clipped) for each of the three univariate functions. The top row shows the non-neural, the bottom row the neural methods.\label{fig:singletrial}}
    \label{fig:example0.1}
\end{figure*}

\begin{table*}[ht]
\caption{Active neurons on univariate tasks when evaluated on the test sets.
For \MM, a neuron was not active in a trial if it never contributed to an output when the network was evaluated on the test data. 
For SMM, a neuron was regarded as not active in a trial if the partial derivatives of the sum of the predictions on the test set w.r.t\ the parameters of the neuron were all zero. For MM, we report the number of active neurons before and after training.
\label{tab:uniactive}}
\begin{center}
\begin{tabular}{@{}lrrrrrrrrr@{}}
\toprule
& \multicolumn{6}{c}{ \MM } &\multicolumn{3}{c}{ \SMM }\\ 
& \multicolumn{3}{c}{ initial } &\multicolumn{3}{c}{ final }\\ 
\cmidrule(lr){2-4}  \cmidrule(lr){5-7}  \cmidrule(lr){8-10} 
& min & mean & max & min & mean & max & min & mean & max \\
\fsq & 1 & 3.4 & 6 & 2 & 3.4 & 5 & 16 & 31.1 & 36 \\
\fsqrt & 2 & 3.9 & 7 & 1 & 2.1 & 4 & 22 & 33.8 & 36 \\
\fsig & 1 & 3.7 & 7 & 2 & 3.0 & 5 & 14 & 29.9 & 36 \\
overall & 1 & 3.7 & 7 & 1 & 2.8 & 5 & 14 & 31.6 & 36 \\
\bottomrule
\end{tabular}
\end{center}
\end{table*}

\begin{table*}[ht]
\caption{Test errors on univariate tasks for \SMM{} different choices of $K$ and initial $\beta$. Values shown are medians over 11 trails. The mean-squared error (MSE) values are multiplied by $10^3$.
We used \cref{eq:gsimple} and \cref{eq:ysimple} without changing the initialization of the weights and biases. As expected the choice of the initial $\beta$ did not have a big effect.
Thus, $\beta$ should be viewed as an additional weight, not as a hyperparameter. 
\label{tab:unihyper}}
\begin{center}
\begin{tabular}{@{}lrrrrrr@{}}
\toprule
  & \multicolumn{5}{c}{ $\ln \beta$ }\\
 & -3 & -2 & -1 & 0 & 1\\
$K$  & \multicolumn{5}{c}{ \fsq }\\
 \cmidrule(lr){2-6}
2 &  0.0293 &  0.0246 &  0.0269 &  0.0183 &  0.0125\\
4 &  0.0255 &  0.0270 &  0.0240 &  0.0109 &  0.0092\\
6 &  0.0243 &  0.0126 &  0.0124 &  0.0087 & \low{ 0.0058}\\
8 &  0.0122 &  0.0131 &  0.0100 &  0.0078 &  0.0062\\
 & \multicolumn{5}{c}{ \fsqrt }\\
 \cmidrule(lr){2-6}
2 &  0.0598 &  0.0599 &  0.0615 &  0.0632 &  0.0711\\
4 &  0.0547 &  0.0262 &  0.0190 &  0.0265 & \low{ 0.0115}\\
6 &  0.0298 &  0.0255 &  0.0211 &  0.0156 &  0.0123\\
8 &  0.0213 &  0.0222 &  0.0165 &  0.0137 &  0.0143\\
 & \multicolumn{5}{c}{ \fsig }\\
 \cmidrule(lr){2-6}
2 &  0.0071 &  0.0048 &  0.0048 & \low{ 0.0036} &  0.0048\\
4 &  0.0044 &  0.0043 &  0.0046 &  0.0064 &  0.0058\\
6 &  0.0041 &  0.0040 &  0.0059 &  0.0057 &  0.0138\\
8 &  0.0040 &  0.0039 &  0.0063 &  0.0098 &  0.0067\\
\bottomrule
\end{tabular}
\end{center}
\end{table*}

\begin{table*}[ht!]
\footnotesize
\caption{Multivariate tasks, training error. The mean-squared error (MSE) values are multiplied by $10^3$.\label{tab:multitrain}}
\begin{center}
\begin{tabular}{@{}lrrrrrrrrr@{}}
\toprule
 & \SMM & \XGs & \XGsval & \XGl & \XGlval& \HLLs & \HLLl & \LMNs & \LMNl \\
\cmidrule(l){2-10} 
$d=2$ & 0.10 & 0.14 & 0.19 & 0.14 & 0.19 & 0.08 & 0.07 & 0.16 & 0.12 \\
$d=4$ & 0.10 & 0.19 & 0.33 & 0.19 & 0.33 & 0.09 & 0.06 & 0.27 & 0.15 \\
$d=6$ & 0.09 & 0.13 & 0.30 & 0.13 & 0.30 & 0.14 & 0.03 & 0.15 & 0.14 \\
 \bottomrule
\end{tabular}
\end{center}
\end{table*}

\begin{table*}[ht!]
\footnotesize
\caption{Multivariate tasks, degrees of freedom of the neural networks and accumulated training times (on an Apple M1 Pro) in seconds for conducting 21 trials with 1000 training steps each.\label{tab:multitraintime}}
\begin{center}
\begin{tabular}{@{}lrrrrrrrrrr@{}}
\toprule
& \multicolumn{2}{c}{ \SMM } &   \multicolumn{2}{c}{ \HLLs } & \multicolumn{2}{c}{ \HLLl } & \multicolumn{2}{c}{ \LMNs } & \multicolumn{2}{c}{ \LMNl } \\
& time (s) & \dofs & time (s) & \dofs& time (s) & \dofs & \dofs& time (s) & \dofs\\
\cmidrule(l){2-11} 
$d=2$ & 9.68  &  109 & 328.87  &  100 & 432.95  &  121 & 9.72  &  105 & 10.22  &  151 \\
$d=4$ & 9.50  &  181 & 293.86  &  81 & 1236.81  &  256 & 10.13  &  171 & 10.46  &  229 \\
$d=6$ & 9.82  &  253 & 235.91  &  64 & 7682.16  &  729 & 10.47  &  253 & 10.92  &  323 \\
 \bottomrule
\end{tabular}
\end{center}
\end{table*}

\begin{figure*}[bh!]
    \centering
    \includegraphics[width=\textwidth]{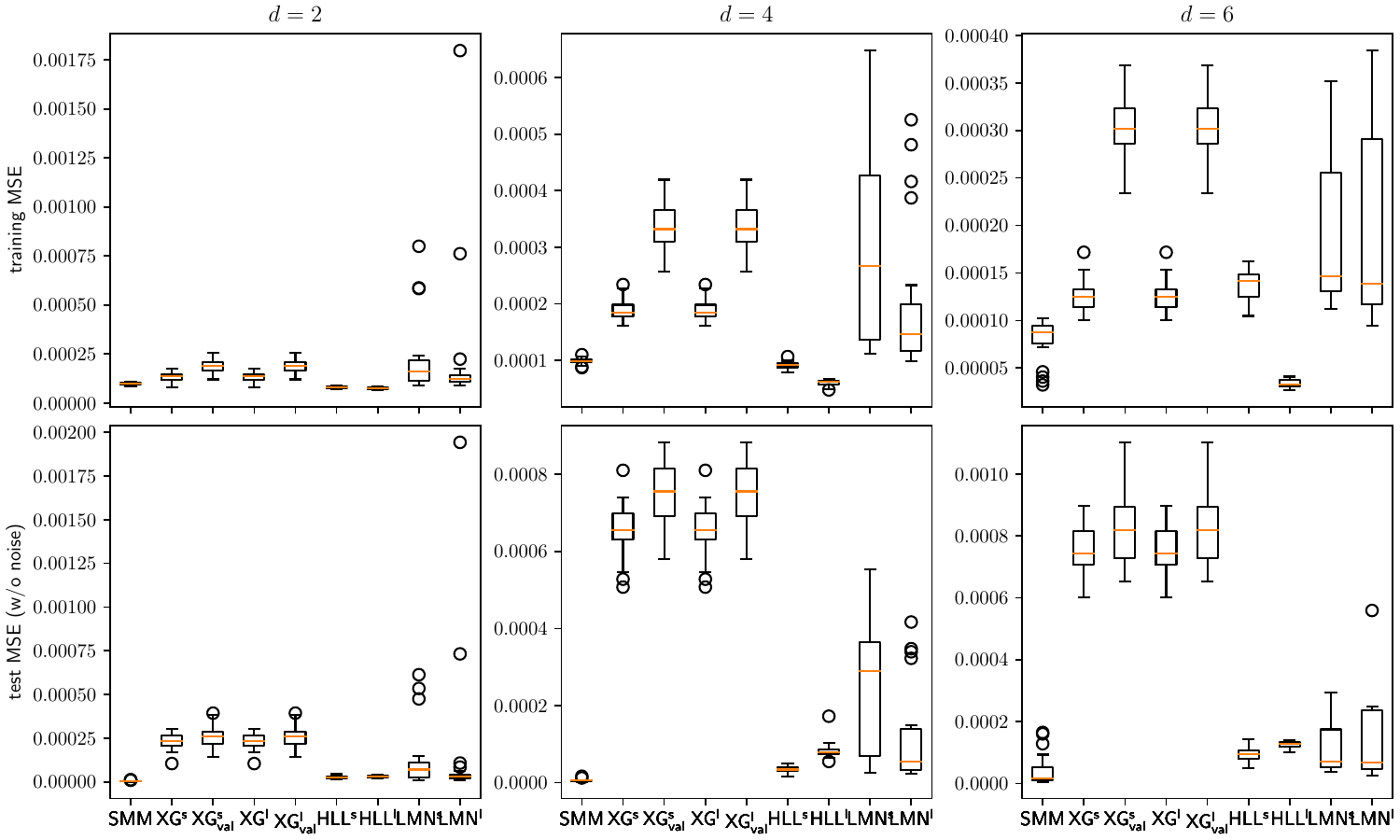}
    \caption{Results on multivariate functions based on $T=21$ trials. Depicted are the median, first and third quartile of the MSE; the  whiskers extend the box by $1\nicefrac{1}{2}$ the inter-quartile range, dots are outliers. Early-stopping reduced the XGBoost training accuracy but did not lead to an improvement on the test data.
    \label{fig:barmulti0.01}}
\end{figure*}

\end{document}